\documentclass{llncs} %

\usepackage{times}
\usepackage{helvet}
\usepackage{courier}
\usepackage[]{todonotes}
\usepackage[english]{babel}
\usepackage{hyperref}
\usepackage{amssymb}
\usepackage{amsmath}
\usepackage{verbatim}

\newcommand{\wdt}[1]{\textsf{\small #1}}

\newcommand{\wdts}[1]{\textsf{\scriptsize #1}}

\newcommand{\tr}[1]{{\color{orange} #1}} 
\newcommand{\trs}[2]{#1 {\color{pink} #2}} 

\renewcommand{\tr}[1]{#1} 
\renewcommand{\trs}[2]{#2}
\renewcommand{\todo}[1]{}

\begin{document}

\trs{
   \title{Wikidata Constraints on MARS}
}{
  \title{Wikidata Constraints on MARS\\(Extended Technical Report)}
}
\author{David Martin \inst{1} \and Peter F. Patel-Schneider \inst{2}} 
\institute{Unaffiliated \and Palo Alto Research Center} 

\maketitle
\begin{abstract}
Wikidata constraints, albeit useful, are represented and processed in
an incomplete, ad hoc fashion. Constraint declarations do not fully
express their meaning, and thus do not provide a precise, unambiguous
basis for constraint specification, or a logical foundation for
constraint-checking implementations.  In prior work we have proposed a
logical framework for Wikidata as a whole, based on multi-attributed
relational structures (MARS) and related logical languages.
In this paper we explain
how constraints are handled in the proposed framework, and
show that nearly all of Wikidata's existing property constraints can
be completely characterized in it, in a natural and economical fashion.
We also give characterizations for several proposed property
constraints, and show that a variety of non-property constraints can
be handled in the same framework.

\end{abstract}

\section{Introduction}

Constraints are extremely useful in Wikidata, as they can be in any
knowledge base.  In Wikidata, property constraints express
regularities (patterns of data) which should hold in general, but 
may have exceptions \cite{PropertyConstraintsPortal}.  In practice,
they are used to identify potential problems (constraint violations)
to interested contributors who can then either fix the problem or
determine that the particular anomaly is acceptable.

One simple example is the \wdt{symmetric constraint}\footnote{For readability we use the
  English label to identify a Wikidata item, here
  \url{https://www.wikidata.org/wiki/Q21510862}.  In formulas, we
  replace spaces with underscores.}
which is understood to indicate that whenever a fact $p(s,
o)$\footnote{We use {\it predicate(subject, object)} notation rather
  than {\it (subject, predicate, object)}.} exists for a symmetric
property $p$ (such as \wdt{spouse}), the fact $p(o, s)$ should normally
also be present.  As of mid-June 2020 there were over thirty-eight
hundred non-symmetric spousal relationships in Wikidata.
We know this because of a report generated by a constraint-checking
tool. Greater contributor effort, or perhaps additional tools, are
needed to determine how many of these non-symmetries are due to
missing \wdt{spouse} statements (as opposed to legitimate exceptions),
and then create them, but that is a separate challenge.  The point
here is simply that this constraint-checking tool has produced a
valuable report.  \todo{Check and reference the tool}


Wikidata constraints, however, are represented and processed in an
incomplete, ad hoc fashion.  Although in most cases they are declared
and documented reasonably clearly, the declarations do not fully
express their meaning.
For example, it is possible to declare that \wdt{spouse} is subject to
the \wdt{symmetric constraint}.
However, crucially, there is no formal characterization of what it
{\it means} for a property to be symmetric.  That is only stated in natural language
documentation.

Stepping outside of Wikidata, it is straightforward to formally
express this meaning in first-order logic (FOL) (with $x$ and $y$ as free
variables, as explained in Section \ref{sec:approach}):
\begin{equation}
\label{eqn:spouse-symmetric-def}
\wdt{spouse}(x, y)  \rightarrow \wdt{spouse}(y, x)
\end{equation}
The value of formal characterizations is foundational in Computer
Science.  We rely on them for clarity in specification in most of our
activities.
And yet Wikidata lacks the logical framework to take advantage of
characterizations like Formula (\ref{eqn:spouse-symmetric-def}).
Such a framework, if available, would provide a precise
basis for constraint specification, and a logical foundation for
constraint-checking implementations.


Further, in current practice specifying a new constraint, and
building a constraint checker for it, may be unnecessarily laborious,
idiosyncratic, and error-prone.  \todo{Should further investigate the
  existing constraint checkers, and extent of sparql usage} A logical formulation and
implementation of constraints would permit constraints to be quickly
specified and reduce the implementation burden for each new type of
constraint.

In prior work \cite{PatelSchneiderWikidataOnMars}, building on the
work of Marx et al. \cite{marx2017logic}, we have proposed a logical
framework for Wikidata, which supports the specification of rules that
can be used to draw inferences to achieve a much more complete
collection of facts, which in turn can support a more comprehensive,
effective, and easy-to-use query service over Wikidata.  This is done
in a way that accounts for, leverages, and facilitates the use of the
representational conventions in Wikidata.

Our logical framework also encompasses the handling of constraints.
In this paper, we describe how this is done, and show that nearly
all of
Wikidata's existing property constraints can be given a complete
characterization in a natural and economical fashion, using a familiar
style of logical expression.  These logical formulae, unlike
documentation in natural language, provide an unambiguous basis for
understanding constraints and for implementing constraint
checkers. (Indeed, once an evaluation capability exists for these
formulae, checking a new constraint requires no
new engineering effort.)  
We also give characterizations for several proposed property constraints that
could usefully be added to Wikidata.
In addition, we show that our approach allows for representing and
handling a broader range of constraints, going beyond property
constraints, in the same formalism.

In the next two sections, we give a general overview of the current
handling of constraints in Wikidata, and an overview of our logical
framework for Wikidata.  In three sections after that, we give examples of
our approach's characterization of existing property constraints,
proposed property constraints, and several useful non-property
constraints.  We follow that with discussion, related work, and
conclusion sections.

\begin {comment}
\subsection{Contributions}
\begin{itemize}
  \item approach
  \item existing property constraints
  \item proposed property constraints
  \item non-property constraints
\end{itemize}
\end{comment}

\section{Property Constraints in Wikidata}

In current Wikidata practice, ``constraint'' is used for both ``property
constraints'' and ``complex constraints''.  We give here an overview
of Wikidata property constraints.  Complex constraints\footnote
{https://www.wikidata.org/wiki/Template:Complex\_constraint} (also
known as ``custom constraints'') are not considered in this paper,
although our approach can handle constraints beyond property constraints.

\begin {comment}
Property constraints are rules on properties that specify how
properties should be
used \cite{PropertyConstraintsPortal}. The
Wikidata model itself is very flexible: nothing stops you from adding
universe (Q1) as head of government (P6). However, a constraint on the
property can tell you that a head of government (P6) is usually a
human.
\end{comment}

At present there are 30 property constraint types used in Wikidata, as
revealed by the ``up-to-date list'' SPARQL query link included on \cite{PropertyConstraintsPortal}.
As explained on that page, ``constraints for a property are specified
as statements on the property, using \wdt{property constraint} (P2302)
and the constraint type item''. For example, in the notation we've adopted for this paper the
following statement says that \wdt{spouse} (P26) is constrained by
the \wdt{symmetric constraint} (Q21510862) constraint type.
\begin{equation*}
\label{eqn:spouse}
\begin{aligned}
  & \wdt{property\_constraint}(\wdt{spouse}, \wdt{symmetric\_constraint})
\end{aligned}
\end{equation*}
\begin {comment}
49 are returned here (URL commented), but some are clearly bogus:


But only 37 are listed here:
https://www.wikidata.org/wiki/Help:Property\_constraints\_portal/list\_of\_constraints
And suggestion constraint is bogus, also wikidata constraint scope, also relation of type constraint, value constraint

And I only see 28 mentioned here:
https://www.wikidata.org/wiki/Wikidata:2020\_report\_on\_Property\_constraints
(Including the 2 ``Lexeme'' constraints, which don't link to any page)

Constraints are hints, not firm restrictions, and are meant as a help
or guidance to the contributor. They can have exceptions: for example,
the town of Talkeetna (Q668224) elected the cat Stubbs (Q7627362) as
mayor.

Some constraint types, currently one-of constraint and allowed
qualifiers constraint, are also used to provide better suggestions
when editing statements.
\end {comment}
Many constraints are configurable by specifying values for
parameters, which are stated as qualifiers on the constraint
statement. (\wdt{Statement} and \wdt{qualifier} in Wikidata are
defined in the Wikibase Data Model \cite{WikibaseDataModel}). There are several
general parameters that can be added to any constraint statement, such
as \wdt{constraint status} (which can have values \wdt{mandatory
  constraint} or \wdt {suggestion constraint}) and \wdt{exception to
  constraint} (which is used to list known exceptions).  There are
other parameters that are specific to a particular constraint type, or
a small group of constraint types.  We shall see examples of some of
these in subsequent sections.

\begin {comment}
  Would like to add more about the following, but out of time and space.
  
Current tools and implementation strategies.
\todo{Delete or enlarge these 2 paragraphs}

Problems / needs (elaborate as needed on points made in the
Introduction).  show that even with the use of sparql, currently, it's
still largely messy and ad hoc.
\end {comment}

\section{Logical Framework}
\label{sec:approach}

Our logical framework for Wikidata \cite{PatelSchneiderWikidataOnMars}
supports the use of both rules and constraints.  Rules are used to
draw inferences; constraints are used to detect the presence of
questionable data patterns.
After briefly reviewing the prior work of Marx et
al. \cite{marx2017logic} -- which produced MARS, MAPL, and MARPL -- we
then introduce our extensions to these -- eMARS, eMAPL, and
eMARPL -- which are the logical foundations of our approach.  In our
approach, rules are expressed in eMARPL, and constraints in eMAPL.

\trs{{\bf MARS, MAPL, and MARPL.}}
{
\subsection{MARS, MAPL, and MARPL}
\label{subsec:marsstuff}}
As noted in \cite{marx2017logic}, Wikidata's custom data model goes
beyond the {\it Property Graph} data model, which associates sets
of attribute-value pairs with the nodes and edges of
a directed graph, by allowing for attributes
with multiple values.  Marx et al.  refer to such generalized
Property Graphs as {\it multi-attributed graphs}, and observes that
``In spite of the huge practical significance of these data models
..., there is practically no support for using such data in knowledge
representation''.  Given that motivation, Marx et al. introduce the
{\it multi-attributed relational structure} (MARS) to provide a formal
data model for generalized Property Graphs, and {\it multi-attributed
  predicate logic} (MAPL) for modeling knowledge over such structures.
MARS and MAPL may be viewed as extensions of FOL to support the use of
attributes (with multiple values).
\tr{In terms of the underlying logical
formalism (which is out of scope here), MARS provides the
structures that serve as interpretations for MAPL.}

The essential new elements over FOL are these:
\begin{itemize}
\item
  a {\it set term} is either a set variable or a set of
  attribute-value pairs $\{a_1 : b_1, ..., a_n : b_n\}$, where $a_i,
    b_i$ can be {\it object terms}.  Object terms are the usual
  basic terms of FOL, and can be either constants or {\it object
    variables}.
  \item
a {\it relational atom} is an expression $p(t_1, ... t_n)@S$, where
$p$ is an {\it n-ary} predicate, $t_1, ... t_n$ are object terms and $S$
is a set term.
\item
  a {\it set atom} is an expression $(a : b) \in S$, where $a, b$ are
  object terms and $S$ a set term.
\end{itemize}

These elements are best illustrated with a simple example (taken directly from \cite{marx2017logic}):

\begin {comment}
\begin{equation}
\label{eqn:spouseRule}
\begin{aligned}
  &\forall x,y,z_1,z_2,z_3 . \\
&\;\;\;\; \wdt{spouse}(x, y)@\{\wdt{start} : z_1, \wdt{loc} : z_2, \wdt{end} : z_3\} \\
&\;\;\;\; \rightarrow \wdt{spouse}(y, x)@\{\wdt{start} : z_1, \wdt{loc} : z_2, \wdt{end} : z_3\}
\end{aligned}
\end{equation}
\end {comment}
\begin{equation}
\label{eqn:spouseRule}
\begin{aligned}
  \forall x,y,z_1,z_2,z_3 . 
\wdt{spouse}(x, y)@\{\wdt{start} : z_1, \wdt{loc} : z_2, \wdt{end} : z_3\} \\
\;\;\;\; \rightarrow \wdt{spouse}(y, x)@\{\wdt{start} : z_1, \wdt{loc} : z_2, \wdt{end} : z_3\}
\end{aligned}
\end{equation}

This MAPL formula states that \wdt{spouse} is a symmetric relation,
where the inverse statement has the same start date, end date, and
location.  Each occurrence of $\wdt{spouse}(...)$ $@\{...\}$ is a relational atom,
which includes the set term $\{\wdt{start} : z_1, \wdt{loc} : z_2,
\wdt{end} : z_3\}$.  If that set term were represented by a set
variable $U$, then one could make an assertion about its membership
using the set atom $(\wdt{start} : z_1) \in U$.

In Wikidata terms, this particular relational atom (once $x$ and $y$
have been instantiated to specific Wikidata items) corresponds to a
statement, and each attribute-value pair (once the $z_i$ variable has
been instantiated to a specific value), corresponds to a qualifier of
the statement.  ($x$, of course, is called the subject of the
statement, and $y$ the value or object of the statement.)  While MAPL
allows for predicates of arbitrary arity, in Wikidata all statements
are triples; i.e. Wikidata properties have arity 2.

Marx et al. go on to introduce multi-attributed rule-based
predicate logic (MARPL), a MAPL fragment which is decidable for fact
entailment, but still provides a high level of expressivity.
\tr{In addition, they define MARPL$_k$, and show that deciding fact
entailment is in polynomial time with respect to data complexity
(i.e., when considering rules, but not data, to be fixed).  Due to
these characteristics of MARPL and MARPL$_k$, along with its logically
well-founded handling of attributes, we believe it to be the best
logical foundation for expressing inference rules in Wikidata, and
supporting reasoning using such rules.}
Note that Formula
\ref{eqn:spouseRule} falls within the MARPL fragment.
MARPL also
allows for a special type of function that can be
used to construct an attribute set in the head of a rule.  A {\it
  MARPL ontology}, then, includes a set of rules and a set of
these function definitions.
Because the representation and checking of
{\it constraints} in our framework builds on MAPL rather than MARPL,
we omit any further details about MARPL.

\trs{{\bf eMARS, eMAPL, and eMARPL.}}
{\subsection{eMARS, eMAPL, and eMARPL}}
MARS / MAPL / MARPL are close to providing a logical basis for
Wikidata but are still missing 2 essential elements:

\begin{itemize}
\item
{\it Wikidata-specific datatypes.} Datatypes play a large role in
Wikidata, and it has its own set of datatypes with certain
idiosyncrasies, as documented in \cite{WikibaseDataModel}.  In order
to specify the manipulation of data elements in rules, functions and
relations are needed for constructing, accessing, and combining the
data elements of each of Wikidata's datatypes.
\item
{\it A feasible means of specifying the uses of attributes in rules.}
Handling Wikidata qualifiers (which are represented as attributes in
MAPL and MARPL) correctly requires accounting for potentially many
attributes in each of many rules, which is infeasible, from a
practical perspective, with MARPL.
\end{itemize}

In \cite{PatelSchneiderWikidataOnMars}, we provide a semi-detailed
sketch for addressing each of these needs.  (A more formal
specification will be provided in a future publication.)
Specifically, we define an {\it extended MARS} (eMARS) as a MARS
extended with a specification of datatypes, with their associated
relations and functions, and we discuss the functions and relations
that are needed for each of Wikidata's datatypes.  We define {\it
  extended MAPL} (eMAPL) to include {\it eMAPL terms}, which are MAPL
terms augmented with datatype function applications, and {\it eMAPL
  formulae}, which allow for the use of eMAPL terms and datatype
relations as predicates.  To further support the representation of
constraints, we also add equality and, as syntactic sugar, counting
quantifiers.

To address the second need mentioned above,
we introduce {\it attribute characterizations}, which provide a means
to describe the desired behavior of attributes when rules fire,
separately from the rules themselves, and we define an {\it extended
  MARPL (eMARPL) ontology} to include, in addition to rules and function definitions,
a set of attribute characterizations.
We also describe how these characterizations can be used as macros,
modifying the functions and rules of an eMARPL ontology.


Given these logical constructs, we show in
\cite{PatelSchneiderWikidataOnMars} how Wikidata itself can be
represented as an eMARS, and discuss some of the essential rules that
are needed for inferencing in Wikidata (including, but not limited to,
ontological rules that axiomatize foundational Wikidata concepts such
as \wdt{instance of}, \wdt{subclass of}, \wdt{subproperty of},
\wdt{reflexive property}, and \wdt{transitive property}).  Other types
of rules are possible and important, such as the rules instantiated in
the SQID tool \cite{marx2017sqid}. The ``meaning of Wikidata'' is
then the inferential closure of the eMARS under an eMARPL ontology
composed of
rules, function definitions, and attribute characterizations.  It is
this eMARS that is used when querying or otherwise requesting what is
true in Wikidata, or checking constraints.

\trs{{\bf Representing Constraints in eMAPL.}}
{\subsection{Representing Constraints in eMAPL}}
We model Wikidata constraints as eMAPL formulae that are evaluated
over the eMARS that is the ``meaning of Wikidata''.  Because
constraint formulae are used as queries, and not for
inferencing, we can take advantage of the greater expressiveness of
eMAPL.  It is known that the data complexity of evaluating
FOL formulae is polynomial, and that remains true for eMAPL formulae.
\todo{reference desirable}

Constraints can either be given a positive formulation, which
expresses a pattern of data elements that conform to the constraint,
or a negative formulation, which expresses a pattern of data elements
that violate the constraint.  In our view, it is most natural to first
write the positive formulation, and from that derive the negative
formulation, which can then be used as a query.  (The derivation of
the negative formulation starts with applying the negation operator to
the positive formulation, and then applies transformations, if
desired, based upon well-known laws of logic.)

For example, the \wdt{distinct\_values\_constraint}\trs{}{\footnote
{https://www.wikidata.org/wiki/Help:Property\_constraints\_portal/Unique\_value}}
in Wikidata indicates that a given property should have different
values for different items (across all of Wikidata).
The following eMAPL formula embodies this constraint.
Here, because we are treating these formulae as queries, the variables are
considered to be free variables.  We omit attribute sets wherever they
are irrelevant to the meaning of the constraint. In other words, for
each atom missing an attribute set there is an implicit variable,
which can be ignored by a constraint-checker (formula evaluator), or treated as an
additional free variable.  \todo {Revisit this with Peter.}
\begin{equation}
\label{eqn:distinctValuesConstraint-a}
\begin{aligned}
\wdt{property\_constraint}(p, \wdt{distinct\_values\_constraint})  \\
\land p(s1, o1) \land p(s2, o2) \land s1 \neq s2 
\rightarrow o1 \neq o2
\end{aligned}
\end{equation}

Formula \ref{eqn:distinctValuesConstraint-a} (the positive
formulation) directly expresses the meaning of the constraint in the
usual fashion of first-order logic.  If satisfied (for all possible bindings of the free variables),
the constraint has no violations.

In all of the
formulae for existing property constraints, we employ Wikidata's
property constraint declarations, which works nicely.  For
example, in Formula \ref{eqn:distinctValuesConstraint-a}, the first
conjunct will match against one of Wikidata's existing property
constraint declarations, thereby binding $p$ to one of the properties
having the distinct values constraint (e.g., the ISBN-13 property, P212).

Formula \ref{eqn:distinctValuesConstraint-b} below (the negative
formulation), where satisfied, identifies items that violate
the constraint.

\begin{equation}
\label{eqn:distinctValuesConstraint-b}
\begin{aligned}
\wdt{property\_constraint}(p, \wdt{distinct\_values\_constraint}) \\
\land \ p(s1, o1) \land p(s2, o2) \land s1 \neq s2 \land o1 = o2
  \end{aligned}
\end{equation}

Because, in our framework, constraints are checked after the KB has
been augmented by running the rules (i.e., the constraints are checked
over the ``meaning of Wikidata'' KB), a far more useful set of results
will be obtained.  Inferences from rules will instantiate facts that
were missing from the original KB, thus providing a complete (with
respect to the rules) set of facts to be checked.  Consequently, a
complete and accurate set of constraint violations will be found, and
false positives and negatives (which would have resulted from missing
facts) will be avoided.

In our framework, as illustrated above, the specification of a new
property constraint type involves, in addition to the creation of
property constraint type declarations of the sort used in current
practice, an eMAPL formula for the new type (or several formulae, if
preferred, in some cases). These formulae, unlike documentation in
natural language, provide an unambiguous basis for understanding and
implementing constraint checkers. Once an evaluation capability exists
for eMAPL formulae, checking a new constraint will require no new
engineering effort.

We investigated the extent to which Wikidata's existing property
constraints can be expressed in eMAPL\trs{, and reported our results
  in \cite{martin2020logical}.  (This paper is a shortened version of
  that report.)}{.}  Out of 26 property constraints examined, only one
could not readily be expressed in eMAPL.  We also became aware of one
{\it proposed} property constraint that cannot readily be
expressed. In both cases, the problem can be addressed in a
straightforward manner.

In the next 2 sections, we show examples of existing and proposed
property constraints, expressed in eMAPL, which illustrate more of its
features. eMAPL allows for representing and handling a broad range of
constraints, going beyond property constraints, in the same
formalism.  In Section \ref{sec:examples}, we illustrate this with
several examples of non-property constraints.  In Section
\trs{\ref{sec:discussion}}{\ref{subsec:limitations}}, we discuss the 2 property constraints that
could not readily be expressed.
\tr{In Section \ref{sec:relatedwork},
we mention some advantages that eMAPL offers over the use of SPARQL.}

\section{Existing Property Constraints}
\label{sec:existing}

In
\trs{\cite{martin2020logical},}{the appendix,}
we give complete characterizations for 26 of the 30
property constraint types in current use.  As explained there, we
omitted 4 constraint types -- the same 4 omitted in
\cite{AbianConstraintsReport} -- due to insufficient documentation
being available for them.
Here, we present two of the 26
characterizations, to illustrate other features of eMAPL.


The {\bf mandatory qualifier constraint
  (Q21510856)}\trs{}{\footnote{https://www.wikidata.org/wiki/Help:Property\_constraints\_portal/Mandatory\_qualifiers}}
  provides a nice illustration of attribute set variables and set
  atoms (from Section \trs{\ref{sec:approach}}{\ref{subsec:marsstuff}}) in the characterization of a constraint type.
Here, we see the set atom $(\wdt{property} : q) \in CQ$ used to obtain
the value $q$ of the \wdt{property} qualifier.  $q$ identifies another
qualifier whose use is mandatory with the given property.  For
example, this constraint type is used with the property
\wdt{population} (P1082).  If this formula were to be evaluated, when
$p$ binds with that property, $q$ will bind with the qualifier
\wdt{point in time} (P585), which is the ``mandatory''
qualifier. $p(s, o)@SQ$ will bind with a fact with property
\wdt{population}, and with statement qualifiers $SQ$. The right-hand
side of the formula, then, checks that $SQ$ contains the mandatory
qualifier.
\begin{equation}
\label{eqn:mandatoryQualifier-pos}
\begin{aligned}
  \wdt{property\_constraint}(p, \wdt{mandatory\_qualifier\_constraint})@CQ \\ 
  \land (\wdt{property} : q) \in CQ  \land 
  p(s, o)@SQ \rightarrow \exists v . (q : v) \in SQ
\end{aligned}
\end{equation}
This is the positive formulation for this constraint type.
\trs{As}{If one
wanted to identify all of the (very many) instantiations that conform
to this constraint type, one could use this positive formulation.  But as}
noted above, in practice one would derive and use the negative formulation to
identify violations\trs{.}{:
\begin{equation}
\label{eqn:mandatoryQualifier-neg}
\begin{aligned}
  \wdt{property\_constraint}(p, \wdt{mandatory\_qualifier\_constraint})@CQ \\ 
  \land (\wdt{property} : q) \in CQ  \land 
  p(s, o)@SQ \land \lnot \exists v . (q : v) \in SQ
\end{aligned}
\end{equation}
} 

The {\bf value type constraint
  (Q21510865)}\trs{}{\footnote{https://www.wikidata.org/wiki/Help:Property\_constraints\_portal/Value\_class}},
which states that each value of the given property should have a given
type (which is also known as the {\it range} of the property) is an
example where it is convenient to express the constraint type with
multiple formulae.  In this case, we use 3 formulae -- one for each
possible value of the \wdt{relation} qualifier (although it could be
done with a single formula if desired).  The \wdt{relation} qualifier
characterizes the allowed relationship between the value and the
type (which is given by the \wdt{class} qualifier). Note also that these
formulae allow for any number of values for the \wdt{class} qualifier,
in keeping with current practice.

\begin{equation*}
\label{eqn:valueType1-pos}
\begin{aligned}
& \wdt{property\_constraint}(p, \wdt{value\_type\_constraint})@CQ  \\
& \land (\wdt{relation} : \wdt{instance\_of}) \in CQ \land p(s, o) \\
& \;\;\; \rightarrow \exists c . ((\wdt{class} : c) \in CQ \land \wdt{instance\_of}(o, c)) \\
& \wdt{property\_constraint}(p, \wdt{value\_type\_constraint})@CQ  \\
& \land (\wdt{relation} : \wdt{subclass\_of}) \in CQ \land p(s, o)  \\
& \;\;\; \rightarrow \exists c . ((\wdt{class} : c) \in CQ \land \wdt{subclass\_of}(o, c)) \\
& \wdt{property\_constraint}(p, \wdt{value\_type\_constraint})@CQ  \\
& \land (\wdt{relation} : \wdt{instance\_or\_ subclass\_of}) \in CQ \land p(s, o)  \\
  & \;\;\; \rightarrow \exists c . ((\wdt{class} : c) \in CQ \land
  (\wdt{instance\_of}(o, c) \lor \wdt{subclass\_of}(o, c)))
\end{aligned}
\end{equation*}
\vspace*{-2ex}

\section{Proposed Property Constraints}
\label{sec:proposed}

Here, we show three other property constraint types that we believe
should be included in Wikidata.  There are many other useful property
constraint types that could be characterized using eMAPL, including
many of the suggested types (determined by survey of active Wikipedia
editors) listed in \cite{AbianConstraintsReport}.

{\bf Asymmetric property constraint.}  Although there is a class \wdt
{asymmetric Wikidata
  property}\trs{}{\footnote{https://www.wikidata.org/wiki/Q18647519}},
\todo{with how many instances} there is no property constraint for
asymmetry.  (This differs from the case of the
class \wdt{symmetric
  property}\trs{}{\footnote{https://www.wikidata.org/wiki/Q18647518}}, which
does have a corresponding property constraint.)  In any case, the
concept of asymmetric property cannot be expressed in eMARPL (and
thus, unlike the case of symmetric property, cannot be expressed as a rule of inference).  
However, asymmetry can easily be expressed as a
constraint in eMAPL, as follows.
\begin{equation}
\label{eqn:asymmetric1-pos}
  \begin{aligned}
  \wdt{asymmetric\_property}(p) \land p(y,x) &\rightarrow \neg p(x,y)
  \end{aligned}
\end{equation}

    {\bf Local value type constraint.}  The concept of a ``local''
    value type constraint has proven to be valuable in ontology
    engineering (where it is sometimes called a ``local range
    restriction'') \todo{Citation would be nice here}, and can easily
    be expressed by extending the characterization of \wdt{value type
      constraint} (see Section \ref{sec:existing}).  ``Local'' in this
    context indicates that the constraint holds when the subject of a
    statement has a particular type, such as the children of humans being humans.
    This constraint can be
    characterized as follows: {\it If the subject item of a statement
      has the given type (indicated using qualifier
      \wdt{local\_class}), the referenced (object) item should be a
      subclass or instance of the given type (indicated using
      qualifier \wdt{class}).}  This constraint calls for a distinct
    \wdt{property constraint} statement for each local class that one
    desires to distinguish for a given property (but it's already
    accepted practice to have multiple \wdt{property constraint}
    statements for a given property and constraint type).
\trs{To save space, here we omit the formulae for the
\wdt{instance\_or\_subclass\_of} and
      \wdt{subclass\_of}  values
      of \wdt{relation}.  }
{
Because we are modeling the declaration of this constraint as an
extension of the \wdt{value type constraint}, we retain the 3 possible
values for the \wdt{relation} qualifier. (We actually have reservations
about the usefulness of the \wdt{subclass\_of} and
\wdt{instance\_or\_subclass\_of} values, not only here but also for
\wdt{type constraint} and \wdt{value type constraint}.  However, a
discussion of their usefulness is out of scope for this paper.)} 
\begin{equation*}
\label{eqn:localValueTypeConstraint1-pos}
\begin{aligned}
& \wdt{property\_constraint}(p, \wdt{local\_value\_type\_constraint})@CQ 
  \land (\wdt{local\_class : lc}) \in CQ \\
& \land (\wdt{relation} : \wdt{instance\_of}) \in CQ
  \land p(s, o) \land \wdt{instance\_of}(s, lc) \\
& \;\;\; \rightarrow \exists c . ((\wdt{class} : c) \in CQ \land \wdt{instance\_of}(o, c))
\end{aligned}
\end{equation*}
\tr{
\begin{equation*}
\label{eqn:localValueTypeConstraint2-pos}
\begin{aligned}
& \wdt{property\_constraint}(p, \wdt{local\_value\_type\_constraint})@CQ
  \land (\wdt{local\_class : lc}) \in CQ \\
& \land (\wdt{relation} : \wdt{subclass\_of}) \in CQ 
  \land p(s, o) \land \wdt{instance\_of}(s, lc) \\
& \;\;\; \rightarrow \exists c . ((\wdt{class} : c) \in CQ \land \wdt{subclass\_of}(o, c))
\end{aligned}
\end{equation*}
\begin{equation*}
\label{eqn:localValueTypeConstraint3-pos}
\begin{aligned}
& \wdt{property\_constraint}(p, \wdt{local\_value\_type\_constraint})@CQ 
  \land (\wdt{local\_class : lc}) \in CQ \\
& \land (\wdt{relation} : \wdt{instance\_or\_subclass\_of}) \in CQ 
  \land p(s, o) \land \wdt{instance\_of}(s, lc) \\
& \;\;\; \rightarrow \exists c . ((\wdt{class} : c) \in CQ \land
  (\wdt{instance\_of}(o, c) \lor \wdt{subclass\_of}(o, c)))
\end{aligned}
\end{equation*}
} 

    {\bf Essential property constraint.}
The importance of a particular property for items of a particular type
could be indicated in a similar fashion to
\wdt{local value type constraint}.  For example, it would be
useful to indicate that a \wdt{person} should normally have a
\wdt{parent} property statement. Because there are persons whose
parents are unknown, a constraint would be more appropriate for this
sort of example than a rule, in our framework.  This
constraint would provide stronger guidance regarding the importance of
a particular property than the existing meta-property \wdt{properties
  for this
  type}\trs{}{\footnote{https://www.wikidata.org/wiki/Property:P1963}}, which
merely indicates the properties that are normally used with items of a
particular type.  Note that the meaning of this constraint is different than that of \wdt
{allowed entity type constraint}, and \wdt{item requires statement
  constraint}.

This property is also ``local'' in the sense that it is conditioned on
the subject of a statement being of a particular type.  In the world
of ontology engineering, this constraint is sometimes called a ``local
existential restriction''.  

\begin{equation}
\label{eqn:essentialPropertyConstraint-pos}
\begin{aligned}
\wdt{property\_constraint}(p, \wdt{essential\_property\_constraint})@CQ \\
\land (\wdt{local\_class : lc}) \in CQ \land \wdt{instance\_of}(s, lc) \rightarrow \exists o . p(s, o)
\end{aligned}
\end{equation}

\section{Non-Property Constraints}
\label{sec:examples}

It is natural to consider a broader range of constraints, and
desirable to express them all in the same logical framework.  Here, we
show eMAPL formulae for several useful constraints that fall outside
the definition of ``property constraint''.  As noted below, some of
these are already present in Wikidata (in some other form besides a
constraint). For those that are already present, we leverage the
existing Wikidata declarations (as we have done for property
constraints). To the best of our knowledge, in current Wikidata
practice these examples would normally be checked by creating a bot,
which would require a greater effort than simply evaluating one of
these formulae (as could be done in our proposed framework), and
the effort would likely be relatively ad hoc, cumbersome, and
error-prone.

\subsection{Union of Classes and Disjoint Classes}

The existing \wdt{union of}\trs{}{\footnote
{https://www.wikidata.org/wiki/Property:P2737}} and \wdt{disjoint union
  of}\trs{}{\footnote {https://www.wikidata.org/wiki/Property:P2738}}
(meta-)properties can each be expressed with a pair of formulae.  Here, we
use the ``dummy value'' \wdt{list\_values\_as\_qualifiers}\trs{ (Q23766486)}{\footnote
{https://www.wikidata.org/wiki/Q23766486}} with \wdt{of}\trs{ (P642)}{\footnote
{https://www.wikidata.org/wiki/Property:P642}}, in accord with existing
practice for these properties.
\vspace*{-1ex}
\begin{equation*}
\label{eqn:unionOf1-pos}
  \begin{aligned}
    & \wdt{union\_of}(u, \wdt{list\_values\_as\_qualifiers})@Q \land \wdt{instance\_of}(i, u)   \\
    & \;\;\; \rightarrow \exists c . ((\wdt{of} : c) \in Q \land \wdt{instance\_of}(i, c)) \\
    & \wdt{union\_of}(u, \wdt{list\_values\_as\_qualifiers})@Q \land
    (\wdt{of} : c) \in Q \land \wdt{instance\_of}(i, c) \\
    & \;\;\; \rightarrow \wdt{instance\_of}(i, u)
  \end{aligned}
\end{equation*}
\vspace*{-2ex}
\begin{equation*}
\label{eqn:disjointUnionOf1-pos}
  \begin{aligned}
   & \wdt{disjoint\_union\_of}(u, \wdt{list\_values\_as\_qualifiers})@Q \land \wdt{instance\_of}(i, u)  \\
   & \;\;\; \rightarrow \exists c1 . ((\wdt{of} : c1) \in Q \land \wdt{instance\_of}(i, c1) \\
   & \;\;\; \;\;\; \;\;\; \land \forall c2 . (((\wdt{of} : c2) \in Q \land \wdt{instance\_of}(i, c2)) \rightarrow c1 = c2)) \\
    & \wdt{disjoint\_union\_of}(u, \wdt{list\_values\_as\_qualifiers})@Q \land
    (\wdt{of} : c) \in Q 
    \land \wdt{instance\_of}(i, c) \\
    & \;\;\; \rightarrow \wdt{instance\_of}(i, u)
  \end{aligned}
\end{equation*}

\vspace*{-0.1ex}
\noindent \wdt{disjoint with}\trs{}{\footnote{https://www.wikidata.org/wiki/Wikidata:Property\_proposal/disjoint\_with}},
a proposed property, was discussed in 2016 but not adopted. In our
opinion, it would be a valuable addition to Wikidata.
\vspace*{-1ex}
\begin{equation*}
\label{eqn:disjointWith-pos}
  \begin{aligned}
    \wdt{disjoint\_with}(c1, c2) &\rightarrow \neg \exists i . (\wdt{instance\_of}(i, c1) \land \wdt{instance\_of}(i, c2))
  \end{aligned}
\end{equation*}
\vspace*{-4ex}

\subsection{No-value Constraint}

We think the best treatment of a no-value snak\footnote
{https://www.mediawiki.org/wiki/Wikibase/DataModel\#PropertyNoValueSnak}
is as a constraint but it is unclear whether a no-value snak means no
value at all, no value with the same qualifiers (as the no-value
snak), or something in between.  These options can be modelled as
eMAPL constraint formulae.  Note that the some-value snak doesn't call
for a constraint, but is addressed by other means in
\cite{PatelSchneiderWikidataOnMars}.

Formula \ref{eqn:noValue-pos} captures the ``no value at all''
interpretation.  Note that \wdt{no\_value}$(p, s)$ statements do not
exist {\it per se} in Wikidata, but could be generated from Wikidata's
internal representation of \wdt{PropertyNoValueSnak}.
\vspace*{-1ex}
\begin{equation}
\label{eqn:noValue-pos}
  \begin{aligned}
\wdt{no\_value}(p, s) \rightarrow \neg \exists o . p(s, o)
  \end{aligned}
\end{equation}

\vspace*{-1ex}
Formula \ref{eqn:noValueSameQualifiers-pos} captures the ``no value with same qualifiers''
interpretation.  
\vspace*{-1ex}
\begin{equation}
\label{eqn:noValueSameQualifiers-pos}
  \begin{aligned}
\wdt{no\_value}(p, s)@Q \rightarrow \neg \exists o . p(s, o)@Q
  \end{aligned}
\end{equation}
\vspace*{-1ex}

\vspace*{-3ex}
\subsection{Other Examples}

Formula \ref{eqn:metasubclassOfConstraint-pos} expresses
the existing \emph{metasubclass of}
relation\trs{}{\footnote{https://www.wikidata.org/wiki/Property:P2445}}
between two metaclasses: instances of the metaclass $m1$ are likely to
be subclasses of classes that are instances of the metaclass $m2$.
\begin{equation}
\label{eqn:metasubclassOfConstraint-pos}
  \begin{aligned}
\wdt{metasubclass\_of}(m1, m2)
\land \ \wdt{instance\_of}(c1, m1) \rightarrow \\ \exists c2 . (\wdt{subclass\_of}(c1, c2) \land \wdt{instance\_of}(c2, m2))
  \end{aligned}
\end{equation}
  
Formula \ref{eqn:notInstanceAndSubclass-pos} states that no item
should be both \wdt{instance of} and \wdt{subclass of} the same
other item.  Formula \ref{eqn:noSubclassLoops-pos} disallows loops in
\emph{subclass of} hierarchies.  Neither of these useful
constraints, to our knowledge, are currently declared or
checked in Wikidata.
\todo{check this}
\begin{equation}
\label{eqn:notInstanceAndSubclass-pos}
  \begin{aligned}
\wdt{instance\_of}(i1, i2) \rightarrow \lnot \wdt{subclass\_of}(i1, i2)
  \end{aligned}
\end{equation}
\vspace*{-3ex}
\begin{equation}
\label{eqn:noSubclassLoops-pos}
  \begin{aligned}
\wdt{subclass\_of}(c1, c2) \land c1 \neq c2 \rightarrow \lnot \wdt{subclass\_of}(c2, c1) 
  \end{aligned}
\end{equation}
\vspace*{-2ex}

\section{Discussion}
\label{sec:discussion}

\trs{}
{\subsection{Rules Versus Constraints}}
In a setting such as our proposed framework, there are some logical
characterizations that can be sensibly used as either rules or
constraints.  For example, the concept of {\it symmetric property},
treated as a property constraint in Wikidata and thus included as a
constraint in this paper, could be used as a rule in our framework, if
one considers that it has no exceptions.  We tend towards this view
ourselves, and in fact, offer a rule for \wdt{symmetric
  property}\trs{}{\footnote {Wikidata includes a class \wdt{symmetric
    property}, but it is deprecated.}}  in
\cite{PatelSchneiderWikidataOnMars}, as well as rules that
characterize the meaning of \wdt{reflexive property}, \wdt{transitive
  property}, \wdt{instance of}, \wdt{subclass of}, and
\wdt{subproperty of}. In our framework, if a logical characterization
is considered to be without exception, and can be expressed in eMARPL,
there is no need to express it as a constraint.  This is because the
reasoning provided by firing the rule will ensure that there are no
exceptions to be found by a constraint formula.

Some constraints (any whose eMAPL formula is also an
eMARPL rule) could be used as rules, as-is.
Given a framework that allows for both rules and constraints, such as
our proposed framework, it isn't necessarily obvious in every case
whether a logical characterization should be treated as a rule or a
constraint.  It can depend not only on logical expressiveness, but
also on intuitions and practices that have developed in the community.
For example, the authors' intuition and experience indicate that the
concept of symmetry is inherent in symmetric properties {\it by
  definition} (as can easily be seen in the case of \wdt{spouse} or \wdt{sibling}), and
thus one needn't and shouldn't allow for exceptions.
Space constraints preclude a full discussion of this question of whether
a rule or constraint usage is more suitable for a given logical
characterization.

In current Wikidata practice, there is evidence of considerable
ambivalence about the extent to which property constraints should
allow for exceptions.  The Help page for property
constraints \cite{PropertyConstraintsPortal}
states that ``Constraints are hints, not firm restrictions, and are
meant as a help or guidance to the editor. They can have
exceptions...''.  At the same time, any constraint can be marked with
a \wdt{constraint status} of \wdt{mandatory}, and 29.2\% of
constraints are characterized in this way, whereas only 4.6\% of
constraints have specified allowed exceptions (using the
\wdt{exception\_to\_constraint}
qualifier) \cite{AbianConstraintsReport}.  Moreover, the
``Wikidata:2020 report on Property constraints''
\cite{AbianConstraintsReport} lists as a goal (i.e., an ``ideal state'') for 21 existing property
constraint types that they should have no exceptions (e.g.,``Goal: No
value type constraint on Wikidata has exceptions.'').

We believe this ambivalence exists, in part, because Wikidata doesn't
currently provide an effective representation of rules (or a mechanism
for deriving inferences from them), \todo{Check this claim} and thus
the existing constraints framework has been forced to accommodate some
things that ought to be rules ({\wdt symmetric property},
etc.).  This provides another strong argument for the adoption of a
framework such as ours.

In our framework, because of their use in reasoning, the
expressiveness of rules necessarily must be more limited than that of
constraint formulae.  Thus, there are a few useful logical
characterizations (e.g., {\it union of}, {\it disjoint union of}, {\it
  disjoint classes}) that one might wish to expresses as rules, but would
not be able to.
In such cases, it would be perfectly reasonable to check them as
constraints. If desired, one could arrange by various means to ensure
that violations of these constraints are not allowed to occur, thus
achieving the effect of a rule, albeit in a somewhat more cumbersome
fashion.


\trs{\medskip}
{\subsection{Limitations}
\label{subsec:limitations}}
We identified one existing constraint (the \wdt {Commons link
  constraint}) and also became aware of one proposed property
constraint (acyclicity) that cannot be readily expressed in eMAPL.
The \wdt{Commons link
  constraint}\trs{}{\footnote{https://www.wikidata.org/wiki/Help:Property\_constraints\_portal/Commons\_link}}
requires knowledge that is not contained in Wikidata.  
However, by adding Wikimedia Commons metadata to Wikidata (one
fact per WC page, giving its name and namespace), this constraint
can be easily expressed.
\trs{Additional details are available in \cite{martin2020logical}.}
  {The appendix contains additional details.}
The proposed {\it acyclic} property constraint, mentioned in
\cite{AbianConstraintsReport}, would check whether a property's usage
has caused a cycle (e.g., A is B's mother, B is C's mother, C is A's
mother), which is outside the expressiveness of an FOL-based logic.
However, because eMAPL is used only as a query language, it could be
extended with property path constructs, like those of SPARQL\trs{ \cite{SPARQL}}{\footnote
{\url{https://www.w3.org/TR/sparql11-query/\#propertypaths}}}, which
would allow for the expression of this proposed constraint.

We have not yet encountered any desirable {\it non-property}
constraints that could not be expressed; however, we have not yet performed a
thorough search for candidate non-property constraints.

\section{Related Work}
\label{sec:relatedwork}

While there isn't space to survey the large literature of
logical frameworks for
knowledge bases, we can highlight relevant work from
several slices of that literature.

\emph{SPARQL.} 
SPARQL is used extensively with Wikidata, via the Wikidata RDF dump, and in some
constraint checking is used in somewhat the same way as we envision for
eMAPL.
Indeed, translation to
SPARQL would be one implementation option for handling constraints
expressed in eMAPL.  SPARQL, of course, supports filters and many other
expressiveness features.  However, as also noted in Section
\ref{sec:discussion}, so far we've only identified one proposed
constraint (acyclicity) that goes beyond the expressiveness of eMAPL---and eMAPL
could be extended in a well-understood manner to allow for this.  SPARQL
also has the advantage of being supported by many existing products.

However, eMAPL provides an attribute set notation for qualifiers,
which is far more natural and readable
than using SPARQL over the complex representation of qualifiers in the
RDF dump.  Similarly, eMAPL provides Wikidata-specific datatype
functions and relations, which, again, results in simpler, more
natural, more compact expressions in some
cases.\trs{}{\footnote{Wikidata-specific datatype functions and relations are
  needed, for example, in the \wdts{contemporary}, \wdts{difference
    within range}, \wdts{format}, \wdts{integer}, and \wdts{range}
  constraints, as shown in \trs{\cite{martin2020logical}}{the appendix}.}}
eMAPL allows for deployment options that are more integral with the
native deployment of Wikidata, thus removing dependency on the RDF dump,
and potentially allowing for more continuous, up-to-date constraint
checking.
At the same time, eMAPL provides a logical foundation for a broader array
of deployment options that are external to Wikidata's native
deployment.

\emph{Constraints in KBs.}
Wikidata's (and our) perspective on constraints is consistent with the
view taken by other recent work on constraints for
knowledge-graph-like systems. The SHACL Shapes Constraint Language
\cite{shacl}, a W3C Recommendation since July 2017, and the Shape
Expressions Language 2.1 (ShEx) \cite {shex} are each used to specify
valid data patterns in RDF KBs, and provide a framework for
identifying violations of those patterns.  The primary differences
from our approach are that they are RDF-specific, and are grounded in
pattern matching techniques rather than in evaluation of logical
formulas.  In addition, our approach provides support for
Wikidata-specific data types and Wikidata's use of qualifiers, and
benefits from its role in a larger logical framework that supports
rule-based inference.
\cite {DBLP:journals/corr/Patel-Schneider14} shows how
Description Logic axioms (when interpreted in a closed-world setting)
can be used for constraint checking, discusses their applicability to
RDF KBs, and shows the feasibility of translation to SPARQL as an
implementation strategy.  The approach herein builds on FOL rather
than Description Logic, and again, addresses challenges specific to Wikidata.

\emph{Logical foundations for Wikidata.}
SQID \cite{marx2017sqid} is a browser and editor for Wikidata, which draws inferences from a collection of MARPL rules.\trs{}{\footnote{SQID's rule set may be viewed at https://tools.wmflabs.org/sqid/\#/rules/browse.}}  Our work was informed by SQID's embodiment of MARPL-based reasoning, and motivated in part by the desire to expand the expressiveness of MARPL rules, as illustrated by the SQID rule set 
to provide a more complete reasoning framework, and to accommodate Wikidata constraints.
 \cite{hanika2019discovering} also formalizes a model of Wikidata based on  MARS, but with a different objective: the application of ``Formal Concept Analysis to efficiently identify comprehensible implications that are implicitly present in the data''.  \cite{hanika2019discovering} is thus nicely complementary with \cite{marx2017logic} and with our work, in that it provides a basis for discovering, rather than hand-authoring, new (e)MARPL rules. 
 


\tr{
\emph{Logical foundations for annotated KBs.} 
Annotated RDFS \cite{zimmermann2012general} extends RDFS and RDFS semantics
to support annotations of triples. A
deductive system is provided, and extensions to the SPARQL query language that enable querying of annotated graphs.  While this approach could provide a useful target formalism for Wikidata's RDF dumps, we have chosen instead to represent Wikidata's data model as directly as possible, and thus we deliberately avoid the use of the RDF dumps, and the complexities that could arise from adopting RDF as the modeling framework.} 

\tr{
\emph{Adding attributes to logics.}
Just as MARPL was developed to provide a (rule-based, Datalog-like) decidable fragment of MAPL, Krötzsch, Ozaki, et al. have also explored description logics as a basis for other
decidable fragments of MAPL, and have analyzed the resulting family of attributed DLs in \cite{krotzsch2017attributed,krotzsch2017reasoning,krotzsch2018attributed,ozaki2018happy,ozaki2019temporally}.  We believe that MARPL provides the best available starting point for modeling Wikidata, but we also agree that this ongoing thread of research will lead to attributed DLs with the right level of expressivity for other sorts of applications.} 

\section{Conclusion and Future Work}

After reviewing our prior work that proposes a logical framework for
Wikidata, based in part on extended multi-attributed predicate logic
(eMAPL), we showed how the framework can be used to give logical
characterizations (eMAPL formulas) for constraints in Wikidata, in a
manner that makes use of Wikidata's existing constraint declarations,
but goes beyond them to give a complete expression of their meaning.
We explained, at a high level, how constraint checking would take
place in our framework.
We are only aware of two property constraints (one existing, one
proposed) which cannot currently be expressed in eMAPL; we explained how
these could be addressed with extensions (to Wikidata content in one
case, eMAPL in the other).  Characterizations are also given for
several proposed property constraints, and for several non-property
constraints whose use could benefit Wikidata.

In future work, we plan to develop a detailed design for a scalable
deployment of our proposed logical framework, in a manner that could
integrate well with existing Wikidata infrastructure, workflow, and
practices.  We also plan to give eMAPL characterizations of the
suggested property constraint types (determined by survey of active
Wikipedia editors) in \cite{AbianConstraintsReport}, and analyze
Wikidata's existing complex constraints and the degree to which they
could be accommodated in our framework.

\bibliographystyle{splncs} 
\bibliography{aaai}

\tr{\section*{Appendix: Existing Property Constraints}
\label{sec:appendix}

Here we present constraint formulae for 26 of the 30 property
constraints in current use (as revealed by the ``up-to-date list''
SPARQL query link included on \cite{PropertyConstraintsPortal}), in
their positive formulations.  These are the same 26 property
constraints covered by the ``Wikidata:2020 report on Property
constraints'' \cite{AbianConstraintsReport}.  We have omitted coverage
for 4 property constraints which are inadequately documented (no Help
pages that we could find); we expect to investigate them in future
work, and do not expect to have any difficulty in characterizing them.

Only one of these 26 constraint characterizations, the
\wdt{Commons link constraint}, requires an
extension to eMAPL, as explained below.  We do not account for uses of
the \wdt{constraint
  scope}\footnote{https://www.wikidata.org/wiki/Property:P4680}
parameter; this will be addressed in future work.

The brief constraint descriptions given here, in
italics, have been adapted from the Wikidata Help page for property
constraints \cite{PropertyConstraintsPortal}, or from the individual
property constraint Help pages linked from there.

Variable $CQ$ abbreviates Constraint Qualifiers; $SQ =$ Statement Qualifiers; $s =$ subject; $p =$ predicate; $o =$ object; $i =$ item or instance; $t =$ type $c =$ class.

\vspace{12pt}

\noindent
    {\bf Commons link constraint (Q21510852)}: {\it Values for the
      property should be valid names of existing pages on Wikimedia
      Commons within a certain specified namespace.}

This property constraint needs access to information from outside of
Wikidata.  To express the constraint in eMAPL requires converting
information in Wikimedia Commons to eMAPL formulae.  A simple
conversion is to create an eMAPL (atomic) formula for each Wikimedia
Commons page that provides the namespace information for the page
name, using the predicate \wdt{Commons\_namespace}.  Then the
constraint is simply a check that the appropriate formula is true.
There is a constraint formula to check that the page exists and one to
check that it is in the correct namespace.

\begin{equation*}
\label{eqn:commonLinkConstraint-pos}
\begin{aligned}
  & \wdt{property\_constraint}(p, \wdt{Commons\_link\_constraint})@CQ  \land p(s, o) \rightarrow \\
  & \;\;\; \exists n \wdt{Commons\_namespace}(o,n) \\
  & \wdt{property\_constraint}(p, \wdt{Commons\_link\_constraint})@CQ \land \\ & (namespace : n) \in CQ \land p(s, o) \rightarrow \\
  & \;\;\; \wdt{Commons\_namespace}(o,n)
\end{aligned}
\end{equation*}

\noindent
    {\bf allowed entity types constraint (Q52004125)}:
    {\it The property may only be used on certain entity type(s)}.

\begin{equation*}
\label{eqn:allowedEntityTypesConstraint-pos}
\begin{aligned}
  & \wdt{property\_constraint}(p, \wdt{allowed\_entity\_types\_constraint})@CQ \land 
  p(s, o) \rightarrow \\
& \;\;\; (\wdt{item\_of\_property\_constraint} : t) \in CQ \land \wdt{instance\_of}(s, t)
\end{aligned}
\end{equation*}

\noindent
    {\bf allowed qualifiers constraint (Q21510851)}:
    {\it Only the given qualifiers may be used with the property}.
Note: it's unnecessary to explicitly mention the special case for ``no
value'', which is present in the documentation for this constraint
type.
\begin{equation*}
\label{eqn:allowedQualifiersConstraint-pos}
\begin{aligned}
& \wdt{property\_constraint}(p, \wdt{allowed\_qualifiers\_constraint})@CQ \land \\
& p(s, o)@SQ \land (q : v) \in SQ \rightarrow \\
& \;\;\; (\wdt{property} : q) \in CQ
\end{aligned}
\end{equation*}

\noindent
    {\bf allowed units constraint (Q21514353)}:
    {\it Values for this statement should only use certain units (or none).}

Note: it's unnecessary to explicitly mention the special case for ``no
value'', which is present in the documentation for this constraint
type.  In our framework, units are simply datatypes, so the logic
allows one to state that $o$ is in the datatype $u$.
\begin{equation*}
\label{eqn:allowedUnitsConstraint-pos}
\begin{aligned}
& \wdt{property\_constraint}(p, \wdt{allowed\_units\_constraint})@CQ \land p(s, o) \rightarrow \\
& \;\;\; \exists u . (\wdt{item\_of\_property\_constraint} : u) \in CQ \land u(o)
\end{aligned}
\end{equation*}

\noindent
    {\bf citation needed constraint (Q54554025)}:
    {\it Statements for the property should have at least one reference.}
\begin{equation*}
\label{eqn:citationNeededConstraint-pos}
\begin{aligned}
  & \wdt{property\_constraint}(p, \wdt{citation\_needed\_constraint})@CQ \land 
  p(s, o)@SQ \rightarrow \\
& \;\;\; \exists r . (\wdt{reference} : r) \in SQ
\end{aligned}
\end{equation*}

\noindent
{\bf conflicts-with constraint (Q21502838)}:
{\it Items using this property should not have a certain other statement.}
\begin{equation*}
\label{eqn:conflictsWithConstraint1-pos}
\begin{aligned}
& \wdt{property\_constraint}(p1, \wdt{conflicts\_with\_constraint})@CQ \ \land \\
& (\wdt{property} : p2) \in CQ \land (\wdt{item\_of\_property\_constraint} : cv) \notin CQ \land
  p1(s, o1) \rightarrow \\
& \;\;\; \lnot \exists o2 . p2(s, o2)
\end{aligned}
\end{equation*}
\begin{equation*}
\label{eqn:conflictsWithConstraint2-pos}
\begin{aligned}
& \wdt{property\_constraint}(p1, \wdt{conflicts\_with\_constraint})@CQ \ \land \\
& (\wdt{property} : p2) \in CQ \land (\wdt{item\_of\_property\_constraint} : cv) \in CQ \land
  p1(s, o1) \rightarrow \\
& \;\;\; \lnot p2(s, cv)
\end{aligned}
\end{equation*}

\noindent
{\bf contemporary constraint (Q25796498)}:
{\it Two entities linked through a property with this constraint must be contemporary, that is, must coexist at some point in history.}

Here, variables $st1/2$ abbreviate start time, and $et1/2$ end
time. \wdt{date\_of\_birth}, \wdt{inception}, \wdt{start\_time},
\wdt{point\_in\_time},
\wdt{dissolved,\_abolished\_or\_demolished\_date},
\wdt{date\_of\_death}, and \wdt{end\_time} are existing Wikidata
qualifiers.  \wdt{less\_than} and \wdt{overlaps} are datatype
relations included in eMAPL \cite{PatelSchneiderWikidataOnMars}.  Note
also that \wdt{less\_than} applies to the main value of a time
interval.
\begin{equation*}
\label{eqn:contemporaryConstraint1-pos}
\begin{aligned}
& \wdt{property\_constraint}(p, \wdt{contemporary\_constraint})@CQ \ \land 
  p(s, o) \rightarrow \\
& \;\;\;\; ((\forall st1 . \lnot ( \wdt{date\_of\_birth}(s, st1) \lor \wdt{inception}(s, st1) \lor \\
& \;\;\;\;\;\;\;\; \wdt{start\_time}(s, st1) \lor \wdt{point\_in\_time}(s, st1) )) \lor \\
& \;\;\;\;\;\; (\forall et2 . \lnot ( \wdt{date\_of\_death}(o, et2) \lor
\wdt{dissolved,\_abolished\_or\_demolished\_date}(o, et2) \lor \\
& \;\;\;\;\;\;\;\; \wdt{end\_time}(o, et2) \lor \wdt{point\_in\_time}(o, et2))) \lor \\
& \;\;\;\;\;\; (\exists st1 \exists et2 \\
& \;\;\;\;\;\;\;\; ( \wdt{less\_than}(st1,et2) \lor \wdt{overlaps}(st1,et2) ) \land \\
& \;\;\;\;\;\;\;\; ( \wdt{date\_of\_birth}(s, st1) \lor \wdt{inception}(s, st1) \lor \\
& \;\;\;\;\;\;\;\;\;\; \wdt{start\_time}(s, st1) \lor \wdt{point\_in\_time}(s, st1) ) \land \\
& \;\;\;\;\;\;\;\; ( \wdt{date\_of\_death}(o, et2) \lor \wdt{dissolved,\_abolished\_or\_demolished\_date}(o, et2) \lor \\
& \;\;\;\;\;\;\;\;\;\; \wdt{end\_time}(o, et2) \lor \wdt{point\_in\_time}(o, et2) ))) \\
& \;\;\;\; \land \\
& \;\;\;\; ((\forall st2 . \lnot ( \wdt{date\_of\_birth}(s, st2) \lor \wdt{inception}(s, st2) \lor \\
& \;\;\;\;\;\;\;\; \wdt{start\_time}(s, st2) \lor \wdt{point\_in\_time}(s, st2) )) \lor \\
& \;\;\;\;\;\; (\forall et1 . \lnot ( \wdt{date\_of\_death}(o, et1) \lor
\wdt{dissolved,\_abolished\_or\_demolished\_date}(o, et1) \lor \\
& \;\;\;\;\;\;\;\; \wdt{end\_time}(o, et1) \lor \wdt{point\_in\_time}(o, et1))) \lor \\
& \;\;\;\;\;\; (\exists st2 \exists et1 \\
& \;\;\;\;\;\;\;\; ( \wdt{less\_than}(st2,et1) \lor \wdt{overlaps}(st2,et1) ) \land \\
& \;\;\;\;\;\;\;\; ( \wdt{date\_of\_birth}(s, st2) \lor \wdt{inception}(s, st2) \lor \\
& \;\;\;\;\;\;\;\;\;\; \wdt{start\_time}(s, st2) \lor \wdt{point\_in\_time}(s, st2) ) \land \\
& \;\;\;\;\;\;\;\; ( \wdt{date\_of\_death(o, et1)} \lor \wdt{dissolved,\_abolished\_or\_demolished\_date}(o, et1) \lor \\
& \;\;\;\;\;\;\;\;\;\; \wdt{end\_time}(o, et1) \lor \wdt{point\_in\_time}(o, et1) )))
\end{aligned}
\end{equation*}

\noindent
    {\bf difference within range constraint (Q21510854)}: {\it The
      difference between the values for two properties should be
      within a certain range or interval. This constraint is available
      for quantity or date properties.}
\begin{equation*}
\label{eqn:differenceWithinRangeConstraint1-pos}
\begin{aligned}
  & \wdt{property\_constraint}(p1, \wdt{difference\_within\_range\_constraint})@CQ \land \\
  & (\wdt{property} : p2) \in CQ \land (\wdt{minimum\_value} : min) \in CQ  \land
  p1(s, o1) \land p2(s, o2) \rightarrow \\
  & \;\;\;\; o1 - o2 \geq min
\end{aligned}
\end{equation*}
\begin{equation*}
\label{eqn:differenceWithinRangeConstraint2-pos}
\begin{aligned}
  & \wdt{property\_constraint}(p1, \wdt{difference\_within\_range\_constraint})@CQ \land \\
  & (\wdt{property} : p2) \in CQ \land (\wdt{maximum\_value} : max) \in CQ  \land
  p1(s, o1) \land p2(s, o2) \rightarrow \\
  & \;\;\;\; o1 - o2 \leq max
\end{aligned}
\end{equation*}

\noindent
    {\bf distinct values constraint (Q21502410)}: {\it Values for this
      property should be unique across all of Wikidata, and no other
      entity should have the same value in a statement for this
      property.}
\begin{equation*}
\label{eqn:distinctValuesConstraint-pos}
\begin{aligned}
\wdt{property\_constraint}(p, \wdt{distinct\_values\_constraint}) \land \\
p(s1, o1) \land p(s2, o2) \land s1 \neq s2 
\rightarrow o1 \neq o2
\end{aligned}
\end{equation*}

\noindent
    {\bf format constraint (Q21502404)}: {\it Values for this
      property should conform to a certain regular expression pattern.}

Here we assume that \wdt{matches\_regex} is a function associated with
the \wdt{StringValue} datatype (a datatype mentioned in
\cite{PatelSchneiderWikidataOnMars} in connection with eMARS).
\begin{equation*}
\label{eqn:formatConstraint-pos}
\begin{aligned}
& \wdt{property\_constraint}(p, \wdt{format\_constraint})@CQ \land \\
& (\wdt{format\_as\_a\_regular\_expression} : regex) \in CQ \land p(s, o) \rightarrow \\
& \;\;\;\; \wdt{matches\_regex}(o, regex)
\end{aligned}
\end{equation*}

\noindent
    {\bf integer constraint (Q52848401)}: {\it Values of this property should have integer type, i.e. a quantity without decimal places. This constraint type should only be used on properties with quantity datatype.}

Here we assume that \wdt{integer} is a function associated with
the \wdt{QuantityValue} datatype (a datatype mentioned in
\cite{PatelSchneiderWikidataOnMars} in connection with eMARS).
\begin{equation*}
\label{eqn:integerConstraint-pos}
\begin{aligned}
\wdt{property\_constraint}(p, \wdt{integer\_constraint})
\land p(s, o) 
\rightarrow \wdt{integer}(o)
\end{aligned}
\end{equation*}

\noindent
{\bf inverse constraint (Q21510855)}: {\it The property has an inverse
  property, and values for the property should have a statement with
  the inverse property pointing back to the original item.}
\begin{equation*}
\label{eqn:inverseConstraint-pos}
\begin{aligned}
  \wdt{property\_constraint}(p1, \wdt{inverse\_constraint})@CQ \ \land \\ 
  (\wdt{property} : p2) \in CQ  \land 
  p1(s, o) \rightarrow p2(o, s)
\end{aligned}
\end{equation*}

\noindent
{\bf item requires statement constraint (Q21503247)}: {\it Items using
  this property should have a certain other statement.}
\begin{equation*}
\label{eqn:itemRequiresStatementConstraint-pos}
\begin{aligned}
  & \wdt{property\_constraint}(p1, \wdt{item\_requires\_statement\_constraint})@CQ \ \land \\ 
  & (\wdt{property} : p2) \in CQ  \land  p1(s, o) \rightarrow  \\
  & \;\;\;\; \exists val . ((\wdt{item\_of\_property\_constraint} : val) \in CQ \land p2(s, val)) \lor \\
  & \;\;\;\; ( (\lnot \exists v . (\wdt{item\_of\_property\_constraint} : v)) \land \exists val . p2(s, val))
\end{aligned}
\end{equation*}

\noindent
{\bf mandatory qualifier constraint (Q21510856)}: {\it The given qualifier is mandatory for this property.}
\begin{equation*}
\label{eqn:mandatoryQualifierConstraint-pos}
\begin{aligned}
  & \wdt{property\_constraint}(p, \wdt{mandatory\_qualifier\_constraint})@CQ \ \land \\ 
  & \;\;\;\; (\wdt{property} : q) \in CQ  \land 
  p(s, o)@SQ \rightarrow \exists v . (q : v) \in SQ
\end{aligned}
\end{equation*}

\noindent
{\bf multi-value constraint (Q21510857)}: {\it Items should have more
  than one statement with this property (or none).}
\begin{equation*}
\label{eqn:multiValueConstraint1-pos}
\begin{aligned}
  & \wdt{property\_constraint}(p, \wdt{multi\_value\_constraint}) \ \land 
  p(s, o1) \\
  & \;\;\;\; \rightarrow \exists o2 . (p(s, o2) \land o1 \neq o2)
\end{aligned}
\end{equation*}
One could easily increase the functionality of this constraint by
adding a \wdt{minimum\_count} parameter, and leveraging eMAPL's counting quantifier to check for a specified minimum number of
statements.  Here, $\exists_{min} o2$ checks that there are at least $min$ instantiations of $o2$ from evaluating the following expression.
\begin{equation*}
\label{eqn:multiValueConstraint2-pos}
\begin{aligned}
  & \wdt{property\_constraint}(p, \wdt{multi\_value\_constraint})@CQ \ \land \\ 
  & (\wdt{minimum\_count} : min) \in CQ \land p(s, o1) \rightarrow \\
  & \;\;\;\; \exists_{min} o2 . p(s, o2)
\end{aligned}
\end{equation*}

\noindent
    {\bf no bounds constraint (Q51723761) }: {\it The value of the
      property should not be used with upper and lower bounds. This
      constraint type should only be used on properties with quantity
      datatype.}

    Here, \wdt{precise} is a relation associated with the \wdt{QuantityValue} datatype.
\begin{equation*}
\label{eqn:noBoundsConstraint-pos}
\begin{aligned}
& \wdt{property\_constraint}(p, \wdt{no\_bounds\_constraint})@CQ
\land p(s, o) \rightarrow \wdt{precise}(o)
\end{aligned}
\end{equation*}

\noindent
{\bf none of constraint (Q52558054) }: {\it The specified values are not allowed for the property.}
\begin{equation*}
\label{eqn:noneOfConstraint-pos}
\begin{aligned}
  & \wdt{property\_constraint}(p, \wdt{none\_of\_constraint})@CQ \ \land \\
  & \;\;\;\; (\wdt{item\_of\_property\_constraint} : v) \in CQ  
  \rightarrow \lnot \exists s . p (s, v)
\end{aligned}
\end{equation*}

\noindent
{\bf one-of constraint (Q21510859)}: {\it Only the specified values are allowed for the property.}
\begin{equation*}
\label{eqn:oneOfConstraint-pos}
\begin{aligned}
  & \wdt{property\_constraint}(p, \wdt{one\_of\_constraint})@CQ \land  p(s, v)
  \rightarrow \\
  & \;\;\;\; (\wdt{item\_of\_property\_constraint} : v) \in CQ  
\end{aligned}
\end{equation*}

\noindent
{\bf property scope constraint (Q53869507)}: {\it The property should
  only be used in one of the specified ways: for the main value of a
  statement, in a reference or as qualifier.}
\begin{equation*}
\label{eqn:propertyScopeConstraint1-pos}
\begin{aligned}
  & \wdt{property\_constraint}(p, \wdt{property\_scope\_constraint})@CQ \land p(s, o)
  \land \\
  & \lnot \wdt{instance\_of}(s, \wdt{wikidata\_reference}) \rightarrow \\
  & \;\;\;\; (\wdt{property\_scope} : \wdt{as\_main\_value}) \in CQ  \\
  & \wdt{property\_constraint}(q, \wdt{property\_scope\_constraint})@CQ \land \\
  & p(s, o)@SQ \land (q : v) \in SQ \rightarrow \\
  & \;\;\;\; (\wdt{property\_scope} : \wdt{as\_qualifiers}) \in CQ \\ 
  & \wdt{property\_constraint}(p, \wdt{property\_scope\_constraint})@CQ \land \\
  & p(s, o) \land \wdt{instance\_of}(s, \wdt{wikidata\_reference}) \rightarrow \\
  & \;\;\;\; (\wdt{property\_scope} : \wdt{as\_references}) \in CQ  
\end{aligned}
\end{equation*}

\noindent
    {\bf range constraint (Q21510860)}: {\it Values for this property should be within a certain range or interval. This constraint is available for quantity or date properties.}
    
The following 2 formulae, using \wdt{minimum\_value} and
\wdt{maximum\_value} qualifiers, are for use with quantity properties.
Two additional formulae, instead using \wdt{minimum\_date} and
\wdt{maximum\_date} qualifiers, are needed for date properties.  (This
conforms with the current declarations and documentation for this
constraint.)
\begin{equation*}
\label{eqn:rangeConstraint1-pos}
\begin{aligned}
  & \wdt{property\_constraint}(p, \wdt{range\_constraint})@CQ \land \\
  & \;\;\;\; (\wdt{minimum\_value} : min) \in CQ  \land p(s, o) \rightarrow o \geq min
\end{aligned}
\end{equation*}
\begin{equation*}
\label{eqn:rangeConstraint2-pos}
\begin{aligned}
  & \wdt{property\_constraint}(p, \wdt{range\_constraint})@CQ \land \\
  & \;\;\;\; (\wdt{maximum\_value} : max) \in CQ  \land p(s, o) \rightarrow o \leq max
\end{aligned}
\end{equation*}

\noindent
{\bf single best value constraint (Q52060874) }: {\it The property
  should have a single ``best'' value for an item. It may have any number of values,
  but exactly one of them (the ``best'' one, by whatever criteria)
  should have preferred rank.}

The first formula states that there should be at least one statement
with preferred rank; the second formula states that if there are
preferred rank statements with 2 different values, they must be
distinguished by different ``separator values'' for the given
separator qualifier.
\begin{equation*}
\label{eqn:singleBestValueConstraint1-pos}
\begin{aligned}
  & \wdt{property\_constraint}(p, \wdt{single\_best\_value\_constraint})@CQ \rightarrow \\
  & \;\;\;\; \exists s,o,SQ . (p(s, o)@SQ \land (rank : preferred) \in SQ)
\end{aligned}
\end{equation*}
\begin{equation*}
\label{eqn:singleBestValueConstraint2-pos}
\begin{aligned}
& \wdt{property\_constraint}(p, \wdt{single\_best\_value\_constraint})@CQ \ \land \\
& p(s, o1)@SQ1 \land p(s, o2)@SQ2 \land o1 \neq o2 \land \\
& (\wdt{rank : preferred}) \in SQ1 \land (\wdt{rank : preferred}) \in SQ2 \rightarrow \\
& \;\;\;\; \exists sep,sepVal1,sepVal2 . \\
& \;\;\;\;\;\;\;\; ((\wdt{separator} : sep) \in CQ \land (sep : sepVal1) \in SQ1 \ \land \\
& \;\;\;\;\;\;\;\;\;\; (sep : sepVal2) \in SQ2 \land sepVal1 \neq sepVal2)
\end{aligned}
\end{equation*}

\noindent
{\bf single value constraint (Q19474404)}: {\it The property generally only has a single value for an item.}
\begin{equation*}
\label{eqn:singleValueConstraint-pos}
\begin{aligned}
& \wdt{property\_constraint}(p, \wdt{single\_value\_constraint})@CQ \ \land \\
& p(s, o1)@SQ1 \land p(s, o2)@SQ2 \rightarrow \\
& \;\;\;\; (o1 = o2 \land SQ1 = SQ2) \ \lor \\
& \;\;\;\; (\wdt{exception\_to\_constraint} : s) \in CQ \ \lor \\
& \;\;\;\; \exists sep,sepVal1,sepVal2 . \\
& \;\;\;\;\;\;\;\;  ((\wdt{separator} : sep) \in CQ \land (sep : sepVal1) \in SQ1 \ \land \\
& \;\;\;\;\;\;\;\;\;\; (sep : sepVal2) \in SQ2 \land sepVal1 \neq sepVal2))
\end{aligned}
\end{equation*}

\noindent
{\bf symmetric constraint (Q21510862)}:
{\it Statements using this property should exist in both directions.}

\begin{equation*}
\label{eqn:symmetric-pos}
\begin{aligned}
\wdt{property\_constraint}(p, \wdt{symmetric\_constraint}) \land
 p(x, y) 
 \rightarrow p(y, x) 
\end{aligned}
\end{equation*}
To also check that the 2 symmetric facts have the same attribute sets,
simply add the use of the attribute set variable SQ:
\begin{equation*}
\label{eqn:symmetric2-pos}
\begin{aligned}
\wdt{property\_constraint}(p, \wdt{symmetric\_constraint}) \land
 p(x, y)@SQ 
 \rightarrow p(y, x)@SQ 
\end{aligned}
\end{equation*}

Note, however, that it may not be desirable to check that the 2
statement qualifier attribute sets are identical, as indicated by the
example in the Motivation section of \wdt{WikiProject Reasoning}
\footnote{https://www.wikidata.org/wiki/Wikidata:WikiProject\_Reasoning}. Given
that, one could introduce a new constraint parameter to specify
exactly which statement qualifiers should be the same (or, conversely,
which statement qualifiers are not required to be the same), and
craft a constraint expression to check that.

\noindent
    {\bf type constraint (Q21503250)}:
    {\it Items with the specified property should have the given type.}

Here we have chosen to have a separate formula for each possible value
of the \wdt{relation} qualifier (although it could be done with a
single formula if desired).  Note also that these formulas allow for
any number of values for the \wdt{class} qualifier, in keeping with
current practice.
\begin{equation*}
\label{eqn:typeConstraint1-pos}
\begin{aligned}
& \wdt{property\_constraint}(p, \wdt{type\_constraint})@CQ \ \land \\
& (\wdt{relation} : \wdt{instance\_of}) \in CQ \land p(s, o) \rightarrow \\
& \;\;\; \exists c . ((\wdt{class} : c) \in CQ \land \wdt{instance\_of}(s, c)) \\
& \wdt{property\_constraint}(p, \wdt{type\_constraint})@CQ \ \land \\
& (\wdt{relation} : \wdt{subclass\_of}) \in CQ \land p(s, o) \rightarrow \\
& \;\;\; \exists c . ((\wdt{class} : c) \in CQ \land \wdt{subclass\_of}(s, c)) \\
& \wdt{property\_constraint}(p, \wdt{type\_constraint})@CQ \ \land \\
& (\wdt{relation} : \wdt{instance\_or\_ subclass\_of}) \in CQ \land p(s, o) \rightarrow \\
  & \;\;\; \exists c . ((\wdt{class} : c) \in CQ \land
  (\wdt{instance\_of}(s, c) \lor \wdt{subclass\_of}(s, c)))
\end{aligned}
\end{equation*}

\noindent
{\bf value requires statement constraint (Q21510864)}: {\it Values for
  this property should have a certain other statement.}

Here, $o$ represents the value (object) of {\it this} property statement,
$val$ the value of the other statement.
\begin{equation*}
\label{eqn:valueRequiresStatementConstraint-pos}
\begin{aligned}
  & \wdt{property\_constraint}(p1, \wdt{value\_requires\_statement\_constraint})@CQ \ \land \\ 
  & (\wdt{property} : p2) \in CQ  \land  
  p1(s, o) \rightarrow  \\
  & \;\;\;\; \exists val . ((\wdt{item\_of\_property\_constraint} : val) \in CQ \land p2(o, val)) \lor \\
  & \;\;\;\; ( (\lnot \exists v . (\wdt{item\_of\_property\_constraint} : v)) \land \exists val . p2(o, val))
\end{aligned}
\end{equation*}

\noindent
{\bf value type constraint (Q21510865)}:
{\it The referenced item should be a subclass or instance of the given type.}

The notes for \wdt{type\_constraint}, above, are applicable here also.

\begin{equation*}
\label{eqn:valueTypeConstraint1-pos}
\begin{aligned}
& \wdt{property\_constraint}(p, \wdt{value\_type\_constraint})@CQ  \\
& \land (\wdt{relation} : \wdt{instance\_of}) \in CQ \land p(s, o) \\
& \;\;\; \rightarrow \exists c . ((\wdt{class} : c) \in CQ \land \wdt{instance\_of}(o, c)) \\
& \wdt{property\_constraint}(p, \wdt{value\_type\_constraint})@CQ  \\
& \land (\wdt{relation} : \wdt{subclass\_of}) \in CQ \land p(s, o)  \\
& \;\;\; \rightarrow \exists c . ((\wdt{class} : c) \in CQ \land \wdt{subclass\_of}(o, c)) \\
& \wdt{property\_constraint}(p, \wdt{value\_type\_constraint})@CQ  \\
& \land (\wdt{relation} : \wdt{instance\_or\_ subclass\_of}) \in CQ \land p(s, o)  \\
  & \;\;\; \rightarrow \exists c . ((\wdt{class} : c) \in CQ \land
  (\wdt{instance\_of}(o, c) \lor \wdt{subclass\_of}(o, c)))
\end{aligned}
\end{equation*}
}

\end{document}